\documentclass{article}

\usepackage[dblblindworkshop, final]{neurips_2026}
\workshoptitle{TAE (Trust-AI-Eval): Can We Trust AI Evaluation?}

\usepackage[utf8]{inputenc}
\usepackage[T1]{fontenc}
\usepackage{hyperref}
\usepackage{url}
\usepackage{booktabs}
\usepackage{amsfonts}
\usepackage{amsmath}
\usepackage{nicefrac}
\usepackage{microtype}
\usepackage{graphicx}

\title{Same Output, Different Gold: Measuring How Reference Choice Moves a Multilingual Benchmark Score}

\author{%
  Parth Kulshreshtha\thanks{Equal contribution.} \\ Centific \\ \texttt{parth.kulshreshtha@centific.com}
  \And
  Shivali Dalmia\footnotemark[1] \\ Centific \\ \texttt{shivali.dalmia@centific.com}
  \And
  Abhishek Mukherji \\ Centific \\ \texttt{abhishek.mukherji@centific.com}
}

\begin{document}

\maketitle

\begin{abstract}
A benchmark score compares a system output against a reference, and
methodological attention falls almost entirely on the first term. We
measure the second. The retained annotation record of a six-language
benchmark for personally identifiable information contains two
independent annotator labellings, the aggregate shipped as gold, and a
reviewer gold from independent expert re-annotation of a sample. Using
it, we hold the scored output fixed and exchange the reference. The score moves by $4.95$ $F_1$ points
for one output and $2.00$ for the other (95\% CIs $[3.23, 6.27]$ and
$[0.26, 3.37]$), and by $7.55$ in the worst language. The
comparison between two outputs moves as well: the paired interaction
between reference and system is $+2.95$ points (CI $[+1.97, +3.87]$),
survives correction for multiple testing, and changes one language's
margin outright. The cause is an undocumented aggregation default that
usually kept one annotator when adjudication did not fire, making the shipped
reference a partial copy of an output being scored. We report the
resulting reference-sensitivity band, show how to compute one from any retained
annotation record, and argue that the quantity belongs beside the score.
\end{abstract}

\section{Introduction}

A benchmark score compares two things: what a system produced, and what a
reference says it should have produced. Almost all methodological
attention falls on the first; the second is treated as a fixed property
of the dataset, reported once and thereafter left implicit. But the
reference is not a constant. It is the output of an annotation process:
people labelled items, they disagreed, and a pipeline resolved those
disagreements according to some rule. A different rule, applied to the
same human labour, yields a different reference. If the score depends
materially on which rule was adopted, then part of what a leaderboard
reports is a property of the annotation pipeline rather than of the
systems it ranks. Reference choice is therefore a source of evaluation uncertainty: a robustness property of the evaluation protocol itself, not only a data-quality concern.

\paragraph{Related work.}
Measures of inter-annotator agreement are long established
\citep{cohen1960coefficient,fleiss1971measuring,krippendorff2019content},
though \citet{artstein2008intercoder} caution that they depend on how
chance agreement is defined. A more recent tradition holds that residual
disagreement is systematic rather than noise
\citep{aroyo2015truth,plank2022problem,uma2021learning}; that work asks
what disagreement means for \emph{training}, whereas we ask what it means
for \emph{scoring}. Closest to us, \citet{northcutt2021pervasive} show
that correcting test-set label errors can reorder model rankings, but they
reconstruct references algorithmically after the fact, and they address
classification, where a label error is a substitution rather than a
differing boundary. \citet{bonagiri2026stableval} find leaderboard
orderings shifting more with the choice of rater than with system
capability, though their references are subsamples of one rater pool, in
English, for preference judgement. A parallel literature asks what scores
license us to claim
\citep{jacobs2021measurement,bowman2021what,raji2021everything}, while
within span extraction the question has been treated as data quality
rather than evaluation sensitivity
\citep{bernier2024annotation,martinez2021single,frenay2014classification}.
For a multilingual span extraction benchmark, we are not aware of prior
work measuring how far a score moves and whether a ranking survives when
the reference is exchanged for an equally defensible alternative from the
same production pipeline.

\paragraph{Our design.}
We invert the usual arrangement: instead of varying the system and fixing
the reference, we hold the system output fixed and vary the reference, so
any movement is attributable to reference choice alone. The measurement
needs a dataset in which the same items carry several independent,
equally defensible references. We use the annotation record of a
production multilingual benchmark for personally identifiable information
(PII) extraction, covering six languages and 24{,}939 prompts. It retains
three things normally discarded: two independent annotator labellings of
nearly every prompt, the aggregate shipped as gold, and an independent
expert re-annotation of a sample. The delivered gold and the expert
re-annotation are both defensible; a practitioner could have adopted
either.

The two references are not interchangeable: the delivered gold
reproduces one annotator's span set more often than the other's in every
one of the six languages, while the expert re-annotation has no such
preference, so the difference is structural rather than sampling noise
(Section~\ref{sec:neutrality}). Exchanging the reference moves the score
by $4.95$ $F_1$ points for one annotator's output (95\% CI
$[3.23, 6.27]$) and $2.00$ for the other (CI $[0.26, 3.37]$), with each
output held fixed. Most consequentially, it changes the comparison
between two outputs: the paired interaction is $+2.95$ points, 95\% CI
$[+1.97, +3.87]$, and survives Holm correction over all 14 tests we
report; the Hindi margin changes sign (Section~\ref{sec:inversion}).

The cause is not exotic. On 94.9\% of audited prompts the shipped
reference is identical to one of the two annotator labellings, because an
undocumented aggregation default usually kept Annotator~A when
adjudication did not fire. The reference was therefore partly copied from the outputs
being compared. We report the resulting movement as a
\emph{reference-sensitivity band} and show how to compute one from any
retained annotation record.

We make three contributions. First, a procedure for measuring reference
sensitivity that requires no new annotation, only an annotation record
that was retained rather than discarded. Second, evidence that an
aggregation default resolving ties by position is an evaluation bias
rather than an annotation detail: it is testable in two lines of code,
and in this pipeline it put a measurable thumb on every score the
benchmark reported. Third, a measurement of what that bias costs on a
six-language span extraction benchmark, with bootstrap intervals
throughout and the limits of the estimate stated.

A note on scope. The outputs we score are human annotation passes, not model predictions. The object we audit, however, is the benchmark evaluation protocol: the same reference-construction process determines the gold against which model outputs would be scored. Human passes are the only outputs this record supports, so whether the magnitude of reference sensitivity transfers to model outputs remains an empirical question. These are measurements of one annotation record, not a general law, and a pipeline with a neutral aggregation rule would show less. That is the point: the size of the effect is a property of the pipeline, and it is currently unreported.

\section{The Dataset and Its Pipeline}
\label{sec:dataset}

We study the annotation record of a production benchmark for PII span
extraction across six languages (Hindi, Portuguese, Polish, Swedish,
Chinese, and Arabic), holding $24{,}939$ prompts after deduplication,
roughly $4{,}100$ per language (Table~\ref{tab:corpus}). The export tags
each prompt with one of seven business domains (finance, IT, insurance,
travel, media, health, retail) and one of twelve task categories, such as
summarisation, translation, and conversation; the median prompt is 418
characters long. A span is a character start offset, an end
offset, and one type, drawn from 21 shared and 77 locale-specific types.

\paragraph{Data and privacy.} Prompts are production customer-service texts used under the data-processing terms of our client agreements, annotated under confidentiality agreements in an access-controlled environment. For the same contractual reasons we release no corpus prompt text, entity value, or annotator identifier, and report only aggregate statistics and type labels.

The record retains up to four labellings per prompt: the redundancy that
makes this study possible, and which most released datasets discard.
\emph{Annotators A and B} labelled each prompt independently, both
present on 99.6\% of prompts; the export carries no annotator identifier,
so all agreement figures are pass-level, not person-level. \emph{The
delivered gold} is what the pipeline shipped as ground truth, and what
any reported score would have been measured against. It is a
\emph{selection}, not a merge: on 94.9\% of audited prompts it is
character-identical to one of the two annotator labellings.
\emph{The reviewer gold} is an independent expert re-annotation from
scratch covering 10.5\% of the corpus. The pipeline's internal
data-preparation record states that reviewed texts were randomly selected
at a 10--15\% rate, and our measured 10.5\% falls inside that range, but
we could not verify the randomness itself from the export. \emph{A selective
third pass}, the pipeline's adjudication step, exists on 35.5\% of
prompts, selected by a rule the export does not record: it covers
65--85\% of prompts on which A and B disagreed, and some on which they
agreed. We use it to explain how the delivered gold was formed but report
no score against it, because a score over a selectively processed subset
would not describe the benchmark.

\begin{table}[t]
\centering
\small
\caption{The annotation record, per language. \emph{Agreement} is micro-$F_1$ between the two annotator labellings over the whole corpus. \emph{Identical} is the share of prompts on which their span sets match exactly. \emph{Type} is the share of boundary-matched spans that also agree on type. \emph{3rd pass} is how often the selective third pass (the pipeline's adjudication step) fired. \emph{Silent} is the share of prompts where the annotators differed, no third pass fired, and the delivered gold reproduced Annotator~A.}
\label{tab:corpus}
\begin{tabular}{lrrrrrrr}
\toprule
 & & \multicolumn{2}{c}{Agreement $F_1$} & & & & \\
\cmidrule(lr){3-4}
Language & Prompts & \textsc{Str.} & \textsc{Rel.} & Identical & Type & 3rd pass & Silent \\
\midrule
Hindi & 4,082 & 73.0 & 76.4 & 50.0 & 95.6 & 61.3 & 9.0 \\
Portuguese & 4,173 & 85.5 & 88.1 & 63.5 & 97.1 & 28.3 & 11.3 \\
Polish & 4,172 & 84.9 & 86.9 & 68.8 & 97.6 & 25.1 & 8.2 \\
Swedish & 4,099 & 72.8 & 75.9 & 48.4 & 95.9 & 38.7 & 14.5 \\
Chinese & 4,213 & 87.8 & 90.8 & 71.8 & 96.7 & 26.4 & 4.2 \\
Arabic & 4,200 & 71.7 & 74.0 & 51.9 & 96.9 & 34.1 & 16.5 \\
\midrule
\textsc{Pooled} & 24,939 & -- & -- & 59.1 & 96.7 & 35.5 & 10.6 \\
\bottomrule
\end{tabular}

\vspace{2pt}
{\footnotesize Pooled agreement $F_1$ is $79.6$ \textsc{Strict} and $82.2$ \textsc{Relaxed}; it is not a mean of the per-language values.}
\end{table}

\subsection{How much the annotators disagree}
\label{sec:ceiling}

Agreement between the two annotator labellings, as micro-averaged $F_1$
on exact span matches, is $79.6$ points strict and $82.2$ relaxed pooled,
ranging from $71.7$ (Arabic) to $87.8$ (Chinese), a $16$-point spread
between languages annotated under the same protocol, by the same vendor,
at the same time (Table~\ref{tab:corpus}).

This agreement is not a ceiling on system performance, since one
inconsistent annotator can lower it without limiting what a careful
system achieves. It is not an estimate of gold accuracy either, but it
varies by $16$ points across languages annotated under one protocol, so
cross-language scores on this benchmark should be read alongside it.

The disagreement is mostly about boundaries. Once both annotators select
the same offsets they assign the same type 95.6--97.6\% of the time, so
they largely agree on the substantive question of whether a string is
PII, and of what kind. What they disagree about is where the entity
begins and ends: whether a title is part of a name, whether a country
code is part of a phone number. We are therefore not arguing that
competent annotators cannot agree on what PII is, but that exact-match
scoring converts a boundary convention into a scoring difference, and
boundary conventions are exactly what a protocol underspecifies. We
report both matching rules throughout so the reader can separate the two
effects. Appendix~\ref{app:examples} gives invented illustrations.

\subsection{Two pipeline mechanisms that shape the gold label}

Two properties of this pipeline determine what the delivered gold
contains. Each is undocumented in any released
description of the benchmark, and each has a measurable downstream
effect.

\paragraph{Disagreement is resolved by default, not by adjudication.}
When the two annotators disagreed and the third pass did not fire, the
pipeline retained Annotator~A in 98.7\% of cases. This occurred on 10.6\%
of prompts pooled, and 16.5\% in Arabic (Table~\ref{tab:corpus},
``Silent''). For those items the gold label is one person's view; the
second annotator recorded a different view, the record preserves it, and
the shipped label does not reflect it. Unlike an annotate-and-adjudicate
protocol, the pipeline did not send every disagreement to adjudication,
and the released labels do not reveal this. Where the third pass did fire on a disagreement, the delivered
gold reproduced B about as often as A ($42.3\%$ and $38.5\%$), so the net
skew toward A that we measure in Section~\ref{sec:neutrality} arises
entirely in the unadjudicated branch.

\paragraph{Adjudication rates are not uniform across languages.}
The third pass fired on 25.1\% of Polish prompts and 61.3\% of Hindi
ones, and Hindi triggers adjudication on roughly 40\% of prompts where
the annotators had already \emph{agreed}, against 3--5\% elsewhere. We
cannot explain this from the export and avoid any cross-language claim
that depends on adjudication behaviour. A benchmark whose adjudication
policy varies by language is not applying one protocol to six languages,
whatever its documentation says. A third mechanism, a type merge applied
before release, is described in Appendix~\ref{app:detail}.

\subsection{The audited subset}
\label{sec:audited}

All four labellings coexist only on the prompts the reviewer sampled. We
call this the \emph{audited subset} and restrict every reference-choice
measurement to it. After deduplication it holds $n = 2{,}628$ prompts
(Hindi 350, Portuguese 630, Polish 339, Swedish 341, Chinese 623, Arabic
345), carrying $9{,}101$ and $9{,}186$ spans from the two annotators. On
it, the reviewer's labelling differs from the delivered gold on 22.0\% of
prompts, from 17.3\% in Portuguese to 29.9\% in Arabic: roughly one gold
label in five does not survive independent expert re-annotation. As
micro-$F_1$ the two references agree at $87.9$--$93.7$ points, differing
from each other by about as much as two annotators do. Restricting to
this subset costs power, which is why we report bootstrap intervals
throughout; what it buys is the only setting in which two equally
defensible references can be compared on identical items with the scored
output held fixed.

\section{Method}

\begin{figure}[t]
\centering
\includegraphics[width=0.85\linewidth]{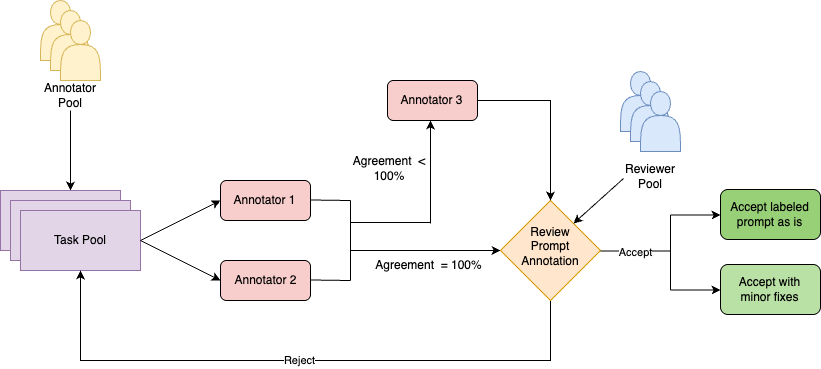}
\caption{The annotation pipeline as designed: two independent passes, a
third on disagreement, and a reviewer pool that re-annotates a sample. In
practice the third pass skipped some disagreements and covered some
agreed prompts (Section~\ref{sec:dataset}). Both references are outputs
of this one process.}
\label{fig:pipeline}
\end{figure}

The design holds one thing fixed and varies exactly one other. The scored
output stays constant; only the reference against which it is scored
changes. The unit of analysis is the prompt, the outputs are the two
annotator labellings, the references are the delivered gold and the
reviewer gold, and the metric is span $F_1$ under two matching rules. All
uncertainty is a paired prompt-level bootstrap, and the ranking claim is
tested as an interaction rather than as two separate verdicts. Figure~\ref{fig:pipeline} shows the pipeline the references come
from. The rest of this section defines each of those in turn: the setting and matching
rules, agreement between labellings, the reference substitution itself,
and uncertainty.

\subsection{Setting and matching rules}

Every measurement is computed on the audited subset of
Section~\ref{sec:audited}, the only items on which all four labellings
coexist. A labelling assigns a set of spans, each a character start
offset, an end offset, and one type; duplicates collapse, and no span
carries two types, so a start--end pair identifies at most one span per
set.

We score under two matching rules. Under \textsc{Strict} a span is the
triple $(\mathit{start}, \mathit{end}, \mathit{type})$ and two spans match
only if all three parts are equal; under \textsc{Relaxed} a span is the
pair $(\mathit{start}, \mathit{end})$ and type is ignored. Both require
exact boundary equality, so \textsc{Relaxed} is type-agnostic
exact-boundary $F_1$ and the gap between the rules isolates the cost of
type disagreement among spans whose boundaries already agree. Neither
credits a partial overlap, so we cannot decompose the remainder into
boundary drift and missed detections, and do not claim to. \textsc{Strict}
is the standard rule for span extraction; we report \textsc{Relaxed}
beside it so the type component can be read off directly.

\subsection{Agreement between two labellings}

We report micro-averaged $F_1$ over span sets, summed over prompts, so
long prompts carry more weight. The measure is symmetric, so we need not
call either side correct, which matters here because which side is
correct is the question under study. We report no chance-corrected
coefficient: annotators choose span boundaries freely, so the number of
chances to agree is undefined \citep{artstein2008intercoder}.
Appendix~\ref{app:defs} gives the definition in full.

\subsection{Reference sensitivity}
\label{sec:refsensitivity}

A benchmark score is a function of the system output and the gold labels.
Standard practice varies the first and fixes the second; we do the
opposite, holding a labelling fixed, treating it as a system output, and
varying which of the others we call gold. We then ask two questions.
First, how far does one score move? We score a fixed labelling against
the delivered gold, then the reviewer gold; the difference is movement
attributable to gold choice alone, since the system did not change.
Second, does the ranking of two outputs survive? We rank the two
annotator labellings under each reference and report whether the ranking
reverses.

One confound needs care. The delivered gold usually equals one of the two
annotator labellings, so scoring an annotator against a gold derived from
it is partly self-scoring and inflates that annotator. We therefore
report two variants: a naive one on the full audited subset, and a
decontaminated one dropping every prompt where the delivered gold equals
either labelling being scored. The naive variant is the main result,
because it is the comparison a practitioner using this benchmark would
actually obtain; the decontaminated variant checks how much of it is
circular. That subset is small ($n = 135$ pooled), so we draw only
negative conclusions from it.

\subsection{Uncertainty}

Every headline number carries a bootstrap percentile confidence interval over
$2{,}000$ resamples with a fixed seed. Resampling is at the prompt level,
because spans inside one prompt are not independent, and the pooled
bootstrap is stratified by language. Every statistic computed on a given
unit reuses that unit's resamples, so all differences and margins are
paired, which is what licenses the interaction test in
Section~\ref{sec:inversion}. A pooled result never stands in for a
per-language one.

\section{Results}

\subsection{The delivered gold is not neutral between annotators}
\label{sec:neutrality}

The two references differ in kind, not merely in content. The delivered
gold reproduces Annotator~A's span set exactly on 83.4\% of prompts and
Annotator~B's on 76.1\%. That is a skew of $+7.3$ points toward A, present
in every language and ranging from $+2.7$ (Polish) to $+12.9$ (Swedish).
The reviewer gold coincides with A on 70.55\% and B on 70.51\%, a pooled
skew of $+0.04$. Per-language reviewer skews range from $-2.3$ (Hindi) to
$+5.3$ (Swedish), so that near-zero is partly a cancellation, with four
of six languages within $1$ point of neutrality. The asymmetry that
matters is nonetheless robust: the delivered gold favours A in
\emph{every} language, the reviewer gold in no consistent direction.
These are not two noisy draws from one distribution: one carries a
systematic preference that is an artefact of the aggregation rule
(Section~\ref{sec:dataset}), the other does not.
Table~\ref{tab:neutrality} gives per-language values.

\subsection{Exchanging the reference moves the score}
\label{sec:goldswap}

Holding each annotator's output fixed and scoring it first against the
delivered gold and then against the reviewer gold, Annotator~A scores
$91.62$ and $86.67$: a difference of $4.95$ points, 95\% CI
$[3.23, 6.27]$. Annotator~B moves by $2.00$ points, CI $[0.26, 3.37]$
(Table~\ref{tab:swapsummary}).
The system did not change between these numbers; only the reference did.

The effect is not uniform. It is largest in Hindi ($7.55$ points, CI
$[4.90, 10.91]$) and Chinese ($7.06$, CI $[4.40, 9.86]$), and reaches
significance in five of six languages for Annotator~A. Across all 24
language--system--rule combinations, 13 deltas exclude zero
(Table~\ref{tab:goldswap}).

Swedish is a clean null: $0.78$ points, CI $[-7.42, 6.58]$.
Section~\ref{sec:neutrality} explains it: Swedish is the one language
whose reviewer gold is itself skewed toward A, by $+5.3$ points. Where
both references share a preference, exchanging them cannot move the
score. The null is consistent with our account rather than a
counterexample, but it bounds the claim: the sensitivity we report is a
property of how these particular references were constructed, not a law
holding wherever two references exist.

\begin{table}[t]
\centering
\small
\caption{Effect of exchanging the reference, \textsc{Strict} matching. Each annotator's output is held fixed and scored against the delivered gold, then the reviewer gold. $\Delta$ is the movement attributable to the reference alone. Bold marks intervals excluding zero. Table~\ref{tab:goldswap} gives all six languages and \textsc{Relaxed}.}
\label{tab:swapsummary}
\begin{tabular}{llrrr}
\toprule
Reference & Ann. & Deliv. & Rev. & $\Delta$ [95\% CI] \\
\midrule
\textsc{Pooled} & A & 91.62 & 86.67 & \textbf{+4.95 [+3.23, +6.27]} \\
 & B & 88.79 & 86.79 & \textbf{+2.00 [+0.26, +3.37]} \\
\midrule
Hindi (largest) & A & 90.29 & 82.74 & \textbf{+7.55 [+4.90, +10.91]} \\
Swedish (smallest) & A & 89.54 & 88.76 & +0.78 [-7.42, +6.58] \\
\bottomrule
\end{tabular}
\end{table}

\subsection{Exchanging the reference changes the comparison}
\label{sec:inversion}

Treating the two annotator labellings as two fixed outputs, the quantity to test is
the \emph{interaction}: the margin between them under the delivered gold
minus the margin under the reviewer gold. Observing that one margin
excludes zero while the other does not is not evidence that the two
differ \citep{gelman2006difference}; the difference itself must be
tested. Both margins are computed inside the same bootstrap resample, so
the interaction is paired and we report it directly.

Pooled, the interaction is $+2.95$ points, 95\% CI $[+1.97, +3.87]$.
Holm-adjusted \citep{holm1979simple} over the fourteen tests in
Table~\ref{tab:inversionmain} it remains significant at $p < 0.01$, as do
Hindi ($+5.27$, CI $[+2.64, +8.05]$) and Portuguese ($+4.42$, CI
$[+2.53, +6.45]$). Swedish falls just outside after correction
($p = 0.08$); Polish, Chinese and Arabic show no interaction
distinguishable from zero. The sign is positive in every language: the
delivered gold consistently favours A more than the reviewer gold does.

\paragraph{The Hindi reversal.}
Hindi is the one language whose margin changes sign. Under the delivered
gold A leads by $3.97$ points, CI $[+0.65, +7.14]$; under the reviewer
gold A trails by $1.30$, CI $[-4.39, +1.55]$. The reversal survives the
relaxed rule and all four duplicate-handling policies. We state its
strength precisely: the reverse leg is not itself significant, so these
data do not show that B is better under the reviewer gold. What is
significant is the interaction, $+5.27$ points with an interval excluding
zero. The defensible claim is that the reference significantly changes
the comparison, and in Hindi changes it far enough to move the sign. We
do not claim a significant reversal; the sign change illustrates the
interaction rather than independently evidencing it.

\paragraph{Multiplicity.}
The pooled interaction is primary; per-language and per-rule breakdowns
are exploratory.

\subsection{How much of this is circular?}
\label{sec:contamination}

The delivered gold is partly built from the outputs being scored, so
scoring A against it is partly self-scoring. This is not a nuisance to be
controlled away. It is the mechanism, and the most transferable finding
in the paper: an aggregation default nobody documented put a measurable
thumb on the scale of every score this benchmark reported.

It also bounds what our numbers support. Repeating the analysis on the
135 prompts where the delivered gold copies neither annotator, A's pooled
advantage moves from $+2.83$ to $-0.09$ points, CI $[-4.52, +4.51]$, and
the interaction falls to $+0.10$, CI $[-2.81, +2.94]$. With $n = 135$ and
13 to 30 items per language every interval straddles zero, and the subset
is far too small to detect an effect of the size we report on the full
data, so it neither confirms nor refutes the interaction. On this
selected subset A's estimated advantage is near zero, as a self-scoring
account predicts, but the interval is too wide to establish whether an
advantage persists after decontamination.
We cannot separate the two effects with these data, and do not claim the
interaction would persist under a neutral aggregation rule.

\subsection{A reference-sensitivity band}
\label{sec:band}

Taking the upper 95\% bound on the largest pooled reference-swap delta gives
$6.3$ $F_1$ points under \textsc{Strict} and $5.2$ under
\textsc{Relaxed}. We call this a \emph{reference-sensitivity band}: the
movement that reference choice alone produces on this benchmark, at the
top of its confidence interval. Three caveats govern its use.

First, it is \emph{descriptive, not an inferential bound}: the upper
interval bound of the largest of 24 measured deltas, selected after
seeing them, with that selection uncorrected. It is neither a minimum
detectable effect nor a significance threshold. The defensible reading is
narrow:
differences smaller than this band are smaller than the movement we
observed from reference choice alone. Second, it is \emph{inflated by self-scoring}. The largest
pooled delta belongs to A, whose score against the delivered gold is
partly circular (Section~\ref{sec:contamination}); the same band computed
from B's delta gives $3.4$ points instead of $6.3$. We report both, and
the gap between them measures how much the aggregation rule contributes.
Third, it is \emph{a property of this pipeline}: one whose aggregation
rule did not favour an annotator would produce a smaller band, possibly a
negligible one, as Swedish demonstrates within our own data.

With those caveats, every margin between the two annotators under the
delivered gold ($2.83$ pooled, magnitudes $1.46$ to $4.59$ across
languages) falls below the $6.3$-point band, and three of seven exceed the stricter
$3.4$-point one. We do not claim the benchmark resolves none of the
comparisons it appears to make. We claim that several of its margins are
the same order as the movement its own reference choice produces, and
that a reader cannot tell which. Every result above survives four
duplicate-handling policies, and Appendix~\ref{app:robustness} reports a
duplicate defect in the record that we excluded throughout.

\section{What This Means for Benchmarks}
\label{sec:discussion}

The standard convention treats a benchmark score as a measurement of a
system. Our results say it measures a system \emph{and} a reference, and
that the second term is not small: larger than every margin
observed in this paper. The second term is invisible not because it
is small, but because the convention reports one number against one
reference and gives it no way to appear.

\subsection{Aggregation rules are evaluation decisions}

The largest driver of our effect is mundane: an implementation default,
not a documented methodological choice, made the shipped reference a
partial copy of one of the things being compared. Any pipeline resolving
ties by position rather than adjudication is open to the same defect. This, rather than the size of any
particular delta, is the transferable finding. The consequences are
systematic: the delivered gold reproduces Annotator~A's span set more
often than B's in all six languages, by $2.7$ to $12.9$ points, so a
benchmark built on it rewards any system whose boundary conventions align
with A's, though neither is more correct. Swedish makes the point from
the other side: where two references share a preference, exchanging them
reveals nothing. That is why sensitivity should be measured per benchmark
rather than assumed.

\subsection{What we recommend}

Four practices follow. \textbf{Retain and release the annotation record},
not only the aggregate. Most pipelines already produce this material and
very few release it, a case \citet{uma2021learning} and
\citet{plank2022problem} make on different grounds.
\textbf{Report a reference-sensitivity band} alongside the headline
score: exchange the reference, hold the outputs fixed, and publish how
far the score and the between-system margin move, testing the
between-system interaction rather than comparing two separate
significance verdicts. What transfers is the procedure, not our number, which is a property of this pipeline. A small
band is a good outcome, not a failed experiment. \textbf{Document
the aggregation rule} and test it for neutrality: how often does the
gold reproduce each annotator exactly? A systematic skew is an evaluation
bias, not an annotation detail. \textbf{Report per-language annotator agreement},
which varies by $16$ points across the six languages here. None requires
new annotation; all require treating the evaluation as an object
of measurement in its own right.

\subsection{Limitations}
\label{sec:limitations}

\textbf{One benchmark, one pipeline.} Every measurement comes from a single annotation record, and the size of the effect depends on properties of that record, in particular an aggregation rule that favours one annotator. We claim the method transfers, not the number. \textbf{The reviewer sample is documented, not observed.} The data-preparation record states the reviewed prompts were randomly selected; we could not verify this from the export. On annotator agreement, the one property we can check, the sample is easier than the corpus in every language (pooled 84.5 vs.\ 79.6 STRICT; Tables~\ref{tab:corpus} and~\ref{tab:matrix}), so selection is more likely to understate the effect than inflate it. \textbf{Pass-level, not person-level.} The export carries no annotator identifier, so we cannot separate a systematic difference in skill from one in convention, and make no claim about which annotator is better. \textbf{The audited subset is small.} $2{,}628$ prompts establish pooled effects but not confident per-language ones, and the decontamination check has $13$--$30$ items per language, so we treat it as inconclusive. \textbf{The reversal is a weakening, not a defeat.} In Hindi the reverse leg is not itself significant; we defend only that a significant win becomes a non-significant loss when the reference is exchanged. \textbf{Two passes are not two systems.} Annotators A and B share guidelines, vendor, and protocol, so their outputs may be more correlated than independently developed systems. Reference sensitivity may therefore differ for model outputs, and we do not infer its magnitude or pattern from these data.

\section{Conclusion}

We held a system output fixed and changed only the reference against
which it was scored. The score moved by $2.00$ to $4.95$ $F_1$ points
pooled, depending on which output was scored, and up to $7.55$ in one
language. The comparison between two outputs moved
too: the paired interaction is $+2.95$ points, 95\% CI $[+1.97, +3.87]$,
and survives correction for multiple testing, while the Hindi margin
changes sign. When two annotators disagreed and no adjudication fired,
the pipeline usually kept the first one. That decision was made once, documented
nowhere, and inherited by every score the benchmark subsequently
reports.

We suggest benchmarks report a reference-sensitivity band alongside the
score. It is a property of the pipeline, not a constant to copy: what
transfers is the procedure. Computing it costs no new annotation, only
keeping the annotation record instead of discarding it, and scoring
against more than one reference. Until reference sensitivity is measured, we cannot know whether a reported improvement is robust to defensible choices in the benchmark's reference-construction process.

\bibliographystyle{plainnat}
\bibliography{references}

\clearpage
\appendix

\section{Full result tables}
\label{app:tables}

The body reports the pooled figures and the extremes. This appendix gives every
per-language value behind them, under both matching rules. Nothing here is new
analysis; it is the same computation reported at full width.

\subsection{Neutrality of each reference}
\label{app:neutrality}

Table~\ref{tab:neutrality} reports, for each reference and each language,
how often that reference reproduces Annotator~A's span set and how often
it reproduces Annotator~B's. A neutral reference should favour neither.
The delivered gold favours~A in all six languages; the reviewer gold
shows no consistent direction. This is the asymmetry that
Section~\ref{sec:neutrality} relies on.

\begin{table}[t]
\centering
\small
\caption{Neutrality of each reference. Percentage of audited prompts on which a reference reproduces an annotator's span set exactly. The delivered gold favours Annotator~A in every language; the reviewer gold does not.}
\label{tab:neutrality}
\begin{tabular}{lrrrcrrr}
\toprule
 & \multicolumn{3}{c}{Delivered gold} & & \multicolumn{3}{c}{Reviewer gold} \\
\cmidrule(lr){2-4}\cmidrule(lr){6-8}
Language & A & B & Skew & & A & B & Skew \\
\midrule
Hindi & 74.6 & 68.6 & +6.0 & & 61.7 & 64.0 & -2.3 \\
Portuguese & 84.6 & 76.5 & +8.1 & & 74.1 & 74.8 & -0.6 \\
Polish & 82.9 & 80.2 & +2.7 & & 72.6 & 73.5 & -0.9 \\
Swedish & 84.2 & 71.3 & +12.9 & & 71.3 & 66.0 & +5.3 \\
Chinese & 90.7 & 83.9 & +6.7 & & 75.8 & 75.6 & +0.2 \\
Arabic & 76.8 & 69.3 & +7.5 & & 60.9 & 61.7 & -0.9 \\
\midrule
\textsc{Pooled} & 83.4 & 76.1 & +7.3 & & 70.5 & 70.5 & +0.0 \\
\bottomrule
\end{tabular}
\end{table}

\subsection{Effect of exchanging the reference, all languages}
\label{app:goldswap}

Table~\ref{tab:goldswap} gives the per-language movement behind the
pooled figures in Section~\ref{sec:goldswap}, for both annotators and
both matching rules, with bootstrap confidence intervals throughout. The
body quotes the pooled value and the extremes; the pattern to read here
is that the effect is positive for Annotator~A in every language but
varies by roughly an order of magnitude in size, and that Swedish alone
is indistinguishable from zero.

\begin{table}[t]
\centering
\small
\caption{Effect of exchanging the reference. Each annotator's output is held fixed and scored against the delivered gold, then the reviewer gold. $\Delta$ is the movement attributable to the reference alone. Bold marks intervals excluding zero.}
\label{tab:goldswap}
\begin{tabular}{llrrrrrr}
\toprule
 & & \multicolumn{3}{c}{\textsc{Strict}} & \multicolumn{3}{c}{\textsc{Relaxed}} \\
\cmidrule(lr){3-5}\cmidrule(lr){6-8}
Language & Ann. & Deliv. & Rev. & $\Delta$ [95\% CI] & Deliv. & Rev. & $\Delta$ [95\% CI] \\
\midrule
Hindi & A & 90.29 & 82.74 & \textbf{+7.55 [+4.90, +10.91]} & 93.17 & 87.31 & \textbf{+5.85 [+3.49, +8.98]} \\
 & B & 86.32 & 84.04 & +2.28 [-0.33, +5.50] & 88.92 & 87.85 & +1.06 [-1.22, +3.99] \\
Portuguese & A & 94.78 & 90.11 & \textbf{+4.67 [+3.31, +6.34]} & 97.12 & 93.26 & \textbf{+3.86 [+2.58, +5.43]} \\
 & B & 90.19 & 89.93 & +0.26 [-1.23, +1.82] & 92.80 & 93.31 & -0.52 [-1.91, +0.94] \\
Polish & A & 89.43 & 85.82 & \textbf{+3.61 [+0.78, +5.92]} & 90.99 & 87.68 & \textbf{+3.31 [+0.54, +5.60]} \\
 & B & 90.89 & 88.64 & +2.25 [-1.16, +5.46] & 91.84 & 89.75 & +2.09 [-1.23, +5.28] \\
Swedish & A & 89.54 & 88.76 & +0.78 [-7.42, +6.58] & 91.41 & 91.41 & -0.01 [-8.22, +5.75] \\
 & B & 85.07 & 87.82 & -2.75 [-10.23, +2.86] & 87.82 & 91.31 & -3.50 [-11.17, +2.16] \\
Chinese & A & 94.34 & 87.28 & \textbf{+7.06 [+4.40, +9.86]} & 95.54 & 89.93 & \textbf{+5.61 [+2.97, +8.40]} \\
 & B & 92.59 & 86.55 & \textbf{+6.04 [+3.68, +8.79]} & 94.50 & 89.87 & \textbf{+4.63 [+2.28, +7.21]} \\
Arabic & A & 86.49 & 80.87 & \textbf{+5.62 [+3.50, +8.12]} & 88.80 & 84.42 & \textbf{+4.38 [+2.40, +6.59]} \\
 & B & 83.29 & 80.12 & \textbf{+3.17 [+0.82, +5.61]} & 84.60 & 82.66 & +1.95 [-0.08, +4.05] \\
\midrule
\textsc{Pooled} & A & 91.62 & 86.67 & \textbf{+4.95 [+3.23, +6.27]} & 93.58 & 89.67 & \textbf{+3.91 [+2.21, +5.18]} \\
 & B & 88.79 & 86.79 & \textbf{+2.00 [+0.26, +3.37]} & 90.86 & 89.78 & +1.08 [-0.70, +2.44] \\
\bottomrule
\end{tabular}
\end{table}

\subsection{Score range available from reference choice}
\label{app:matrix}

Table~\ref{tab:matrix} widens the comparison from the two curated
references to every labelling available in the record, which bounds how
far the score can be moved by reference choice alone. Widening from two curated references to every available labelling,
Annotator~A's pooled \textsc{Strict} $F_1$ ranges from $84.54$ to
$91.62$, a spread of $7.08$ points, reaching $12.46$ in Arabic
(Table~\ref{tab:matrix}). Restricting to the two curated references (the
comparison a practitioner would actually face), the pooled spread is
$4.95$ points. In 13 of the 14 cells the maximum is obtained against the
delivered gold, which follows from Section~\ref{sec:neutrality}; the exception is Swedish~B, consistent with Swedish being the one language whose reviewer gold is also skewed toward A.

\begin{table}[t]
\centering
\small
\caption{Score range available from the choice of reference alone, \textsc{Strict} matching. Each row holds one annotator's output fixed and varies which other labelling is designated gold. The maximum is obtained against the delivered gold in 13 of the 14 rows; the exception is Swedish~B.}
\label{tab:matrix}
\begin{tabular}{llrrrr}
\toprule
Language & Ann. & vs.\ other ann. & vs.\ deliv. & vs.\ rev. & Range \\
\midrule
Hindi & A & 79.69 & 90.29 & 82.74 & \textbf{10.60} \\
 & B & 79.69 & 86.32 & 84.04 & \textbf{6.63} \\
Portuguese & A & 86.06 & 94.78 & 90.11 & \textbf{8.73} \\
 & B & 86.06 & 90.19 & 89.93 & \textbf{4.13} \\
Polish & A & 86.45 & 89.43 & 85.82 & \textbf{3.61} \\
 & B & 86.45 & 90.89 & 88.64 & \textbf{4.44} \\
Swedish & A & 84.25 & 89.54 & 88.76 & \textbf{5.29} \\
 & B & 84.25 & 85.07 & 87.82 & \textbf{3.57} \\
Chinese & A & 90.00 & 94.34 & 87.28 & \textbf{7.06} \\
 & B & 90.00 & 92.59 & 86.55 & \textbf{6.04} \\
Arabic & A & 74.03 & 86.49 & 80.87 & \textbf{12.46} \\
 & B & 74.03 & 83.29 & 80.12 & \textbf{9.26} \\
\midrule
\textsc{Pooled} & A & 84.54 & 91.62 & 86.67 & \textbf{7.08} \\
 & B & 84.54 & 88.79 & 86.79 & \textbf{4.25} \\
\bottomrule
\end{tabular}
\end{table}

\subsection{Ranking survival, \textsc{strict} matching}
\label{app:inversionmain}

Table~\ref{tab:inversionmain} is the summary referenced from
Section~\ref{sec:inversion}: per-language margins under each reference,
the paired swing, and Holm-adjusted $p$-values, under \textsc{Strict}
matching only.

\begin{table}[t]
\centering
\small
\caption{Does the ranking of the two annotators survive a change of reference? Margin is $F_1(\text{A}) - F_1(\text{B})$ under each reference. \emph{Swing} is the paired interaction, (margin under delivered gold) $-$ (margin under reviewer gold), recomputed inside every bootstrap resample; it tests directly whether the reference changes the comparison. Bold marks intervals excluding zero. $p$ is a two-sided bootstrap $p$-value for the swing, Holm-adjusted over all 14 tests. \textsc{Strict} matching; Table~\ref{tab:inversion} adds the \textsc{relaxed} rows, which are included in the Holm family above.}
\label{tab:inversionmain}
\begin{tabular}{llrrrr}
\toprule
 & & \multicolumn{2}{c}{Margin [95\% CI]} & & \\
\cmidrule(lr){3-4}
Language & Rule & Delivered gold & Reviewer gold & Swing [95\% CI] & Holm $p$ \\
\midrule
Hindi & \textsc{strict} & \textbf{+3.97 [+0.65, +7.14]} & -1.30 [-4.39, +1.55] & \textbf{+5.27 [+2.64, +8.05]} & $<$0.01 \\
Portuguese & \textsc{strict} & \textbf{+4.59 [+2.41, +6.71]} & +0.18 [-1.75, +2.12] & \textbf{+4.42 [+2.53, +6.45]} & $<$0.01 \\
Polish & \textsc{strict} & -1.46 [-4.59, +2.34] & \textbf{-2.82 [-5.28, -0.29]} & +1.37 [-0.67, +3.63] & 0.73 \\
Swedish & \textsc{strict} & \textbf{+4.47 [+1.55, +7.68]} & +0.94 [-1.66, +3.75] & \textbf{+3.53 [+0.80, +6.58]} & 0.08 \\
Chinese & \textsc{strict} & +1.75 [-0.54, +3.97] & +0.73 [-1.04, +2.50] & +1.02 [-0.87, +2.80] & 0.73 \\
Arabic & \textsc{strict} & +3.20 [-1.55, +8.03] & +0.75 [-3.64, +5.07] & +2.45 [-0.17, +5.38] & 0.42 \\
\midrule
\textsc{Pooled} & \textsc{strict} & \textbf{+2.83 [+1.57, +4.01]} & -0.12 [-1.15, +0.92] & \textbf{+2.95 [+1.97, +3.87]} & $<$0.01 \\
\bottomrule
\end{tabular}
\end{table}

\subsection{Ranking survival, both matching rules}

Table~\ref{tab:inversion} repeats Table~\ref{tab:inversionmain} under both matching rules. No language changes verdict between them: Hindi and Portuguese swing significantly under \textsc{Strict} and \textsc{Relaxed} alike, the other four are indistinguishable from zero under both, and the Hindi sign change reported in Section~\ref{sec:inversion} is present under each rule.

\begin{table}[t]
\centering
\small
\caption{Does the ranking of the two annotators survive a change of reference? Margin is $F_1(\text{A}) - F_1(\text{B})$ under each reference. \emph{Swing} is the paired interaction, (margin under delivered gold) $-$ (margin under reviewer gold), recomputed inside every bootstrap resample; it tests directly whether the reference changes the comparison. Bold marks intervals excluding zero. $p$ is a two-sided bootstrap $p$-value for the swing, Holm-adjusted over all 14 tests. Both matching rules.}
\label{tab:inversion}
\begin{tabular}{llrrrr}
\toprule
 & & \multicolumn{2}{c}{Margin [95\% CI]} & & \\
\cmidrule(lr){3-4}
Language & Rule & Delivered gold & Reviewer gold & Swing [95\% CI] & Holm $p$ \\
\midrule
Hindi & \textsc{strict} & \textbf{+3.97 [+0.65, +7.14]} & -1.30 [-4.39, +1.55] & \textbf{+5.27 [+2.64, +8.05]} & $<$0.01 \\
 & \textsc{relaxed} & \textbf{+4.25 [+1.28, +7.04]} & -0.54 [-3.23, +1.90] & \textbf{+4.79 [+2.25, +7.37]} & $<$0.01 \\
Portuguese & \textsc{strict} & \textbf{+4.59 [+2.41, +6.71]} & +0.18 [-1.75, +2.12] & \textbf{+4.42 [+2.53, +6.45]} & $<$0.01 \\
 & \textsc{relaxed} & \textbf{+4.33 [+2.57, +6.19]} & -0.05 [-1.62, +1.58] & \textbf{+4.37 [+2.57, +6.41]} & $<$0.01 \\
Polish & \textsc{strict} & -1.46 [-4.59, +2.34] & \textbf{-2.82 [-5.28, -0.29]} & +1.37 [-0.67, +3.63] & 0.73 \\
 & \textsc{relaxed} & -0.85 [-3.85, +2.48] & -2.07 [-4.33, +0.22] & +1.21 [-0.78, +3.45] & 0.73 \\
Swedish & \textsc{strict} & \textbf{+4.47 [+1.55, +7.68]} & +0.94 [-1.66, +3.75] & \textbf{+3.53 [+0.80, +6.58]} & 0.08 \\
 & \textsc{relaxed} & \textbf{+3.59 [+1.13, +6.29]} & +0.10 [-2.16, +2.34] & \textbf{+3.49 [+0.93, +6.58]} & 0.06 \\
Chinese & \textsc{strict} & +1.75 [-0.54, +3.97] & +0.73 [-1.04, +2.50] & +1.02 [-0.87, +2.80] & 0.73 \\
 & \textsc{relaxed} & +1.04 [-0.90, +2.91] & +0.06 [-1.34, +1.56] & +0.97 [-0.87, +2.64] & 0.73 \\
Arabic & \textsc{strict} & +3.20 [-1.55, +8.03] & +0.75 [-3.64, +5.07] & +2.45 [-0.17, +5.38] & 0.42 \\
 & \textsc{relaxed} & +4.20 [-0.41, +8.93] & +1.76 [-2.69, +6.21] & +2.43 [-0.22, +5.27] & 0.42 \\
\midrule
\textsc{Pooled} & \textsc{strict} & \textbf{+2.83 [+1.57, +4.01]} & -0.12 [-1.15, +0.92] & \textbf{+2.95 [+1.97, +3.87]} & $<$0.01 \\
 & \textsc{relaxed} & \textbf{+2.72 [+1.60, +3.78]} & -0.12 [-1.03, +0.83] & \textbf{+2.84 [+1.91, +3.72]} & $<$0.01 \\
\bottomrule
\end{tabular}
\end{table}

\section{Two scoring conventions that void a metric}
\label{app:conventions}

Two further mechanisms attenuate what this benchmark can observe, and
both are invisible in its outputs. The type merge described in
Appendix~\ref{app:detail} collapses types annotators are known to confuse
into one class, after which the confusion cannot be measured and the
benchmark reports agreement it did not obtain. Separately, the original
evaluation located predicted spans by exact string search, under which a
partial overlap is nearly unreachable: across 24 reported result blocks
the ``partial match'' and exact-match metrics are numerically identical
in every one. A benchmark reporting four matching rules was effectively
reporting three, and nothing in the output signals it: it is visible
only by reading the scoring code. We highlight this not as a criticism of
one implementation but as a general failure mode: metric definitions are
realised in code, the code is rarely published alongside the score, and a
metric can be void without any visible symptom.

\section{Additional result detail}
\label{app:detail}

Annotator~A holds a lead excluding zero under the delivered gold in
Hindi, Portuguese, Swedish and pooled. None survives the exchange of
reference (Portuguese $+4.59 \rightarrow +0.18$; Swedish
$+4.47 \rightarrow +0.94$; pooled $+2.83 \rightarrow -0.12$). We report
this descriptively; the inference is the interaction above. Two negative
findings belong here: Polish never reverses, and B leads under both
references; and the pooled and Portuguese-relaxed sign changes rest on
reviewer-gold margins of $-0.12$ and $-0.05$ points, which are ties, not
reversals, and we do not report them as inversions.

\paragraph{A type merge conceals a known confusion.}
Before release, a merge map collapses 25 fine-grained PII types into 17,
placing national identity numbers, tax identifiers, and permanent account
numbers into one \textsc{tax number} class. An earlier internal analysis
of the same pipeline records confusion between two of those types as a
recurring annotator error. The merge does not resolve that error, it
makes it unobservable: after the merge the two labels are the same label,
and the annotators appear to agree. The released type vocabulary is not
the vocabulary the annotators worked in, and the difference between them
absorbs a known source of disagreement.

\section{Robustness checks and the duplicate defect}
\label{app:robustness}

We repeated every analysis under four duplicate-handling policies: drop
all duplicates, keep the first occurrence, keep the last, and retain
every copy. The Hindi reversal reported in Section~\ref{sec:inversion} is
present in all four.

We also identified 11 duplicated items, 10 in one Chinese locale and 1 in
Arabic, whose reviewer annotation is a verbatim copy of one annotator's
span set with zero overlap with the other's. Retaining either copy
manufactures a winner by construction, and doing so produced a spurious
Chinese reversal under one policy. We exclude the 22 affected rows,
$0.8\%$ of the audited subset, from all reported results, and found no
comparable defect elsewhere in the record.

\section{Definitions in full}
\label{app:defs}

Let $A$ and $B$ be the span sets two labellings assign to a prompt. We
report micro-averaged $F_1$, summed over prompts:

\begin{equation}
F_1 = \frac{2 \sum_i |A_i \cap B_i|}{\sum_i |A_i| + \sum_i |B_i|}
\end{equation}

The sums run over all prompts $i$ in the subset, so this is not a
per-prompt average and long prompts carry more weight, matching how span
extraction benchmarks are scored. The measure is symmetric, so we need
not call either side correct, which matters here because which side is
correct is the question under study. We report no chance-corrected
coefficient: span detection has no fixed set of items raters could have
agreed on, since annotators choose boundaries freely in continuous text,
so the number of chances to agree is undefined
\citep{artstein2008intercoder}.

Every headline number carries a bootstrap confidence interval. We
resample prompts with replacement, $2{,}000$ times, with a fixed seed.
Resampling is at the prompt level, not the span level, because spans
inside one prompt are not independent. The pooled bootstrap is stratified
by language, so the language mix is held fixed across resamples. Every
statistic computed on a given unit reuses that unit's resamples, so all
differences and margins we report are paired. We report 95\% intervals
for each $F_1$, each reference-choice difference, and each ranking margin, and
state plainly which differences survive their interval. A pooled result
never stands in for a per-language result.

\section{Invented illustrations of boundary disagreement}
\label{app:examples}

Table~\ref{tab:examples} gives invented prompts that illustrate the kind
of boundary disagreement described in Section~\ref{sec:ceiling}; none
reproduces corpus text or values, and they are not a sample of the
corpus.

\begin{table}[ht]
\centering
\small
\caption{Invented prompts illustrating boundary disagreement. Each row shows only the disputed span; the other spans in the prompt are identical under both labellings. The A/B assignment is illustrative and makes no claim about which annotator did what. All names, numbers, and addresses are fictitious.}
\label{tab:examples}
\begin{tabular}{p{1.9cm}p{11.2cm}}
\toprule
Domain, task & Invented prompt and disputed span \\
\midrule
Finance, Rewrite & Make this note more formal: ``Dr.~Anna Berg phoned about the card ending 0000, please call her back.'' \newline
A: \texttt{Dr.~Anna Berg} \quad B: \texttt{Anna Berg} \quad (NAME) \\
\addlinespace
Travel, Extract & List the contact details in this booking: guest Omar Haddad, omar.haddad@example.com, +46 70 000 00 00. \newline
A: \texttt{+46 70 000 00 00} \quad B: \texttt{70 000 00 00} \quad (PHONE) \\
\addlinespace
Insurance, Summarize & Summarise this claim: the policyholder reports water damage at Flat 4B, 12 Harbour Road, after a pipe burst on 3 March. \newline
A: \texttt{12 Harbour Road} \quad B: \texttt{Flat 4B, 12 Harbour Road} \quad (ADDRESS) \\
\bottomrule
\end{tabular}
\end{table}

\end{document}